\documentclass[11pt,a4paper]{article}
\usepackage[hyperref]{acl2021}
\usepackage{times}
\usepackage{latexsym}
\usepackage{booktabs}
\usepackage{graphicx}

\usepackage{url}
\usepackage{graphicx}
\usepackage{amsmath}
\usepackage{amssymb}
\usepackage{subfigure}
\usepackage{caption}
\usepackage{algorithm}
\usepackage{algorithmicx}
\usepackage{algpseudocode}
\usepackage{multicol}
\usepackage{multirow}
\usepackage{booktabs}
\usepackage{listings}
\usepackage{array}
\usepackage{bm} 
\usepackage{comment}

\usepackage{microtype}

\aclfinalcopy 

\usepackage{xcolor, soul}
	\definecolor{zzycolor}{rgb}{0.688, 0.588, 0.188}

\title{Better Robustness by More Coverage: \\ Adversarial and Mixup Data Augmentation for Robust Finetuning}

  \author{Chenglei Si$^{1,3*}$, Zhengyan Zhang$^{2,3*}$, Fanchao Qi$^{2,3}$, Zhiyuan Liu$^{2,3,4\dagger}$, \\
 \textbf{Yasheng Wang}$^5$, \textbf{Qun Liu}$^5$, \textbf{Maosong Sun}$^{2,3,4}$\\
        $^1$University of Maryland, College Park, MD, USA \\
        $^2$Department of Computer Science and Technology, Tsinghua University, Beijing, China  \\
         $^3$Beijing National Research Center for Information Science and Technology\\
         $^4$Institute for Artificial Intelligence, Tsinghua University, Beijing, China \\ 
         $^5$Huawei Noah's Ark Lab \\
         \texttt{clsi@terpmail.umd.edu, zy-z19@mails.tsinghua.edu.cn}\\
}

\date{}

\begin{document}
\maketitle
\begin{abstract}
Pretrained language models (PLMs) perform poorly under adversarial attacks. To improve the adversarial robustness, adversarial data augmentation (ADA) has been widely adopted to cover more search space of adversarial attacks by adding textual adversarial examples during training.
However, the number of adversarial examples for text augmentation is still extremely insufficient due to the exponentially large attack search space.
In this work, we propose a simple and effective method to cover a much larger proportion of the attack search space, called Adversarial and Mixup  Data Augmentation (AMDA). 
Specifically, AMDA linearly interpolates the representations of pairs of training samples to form new virtual samples, which are more abundant and diverse than the discrete text adversarial examples in conventional ADA.
Moreover, to fairly evaluate the robustness of different models, we adopt a challenging evaluation setup, which generates a new set of adversarial examples targeting each model. 
In text classification experiments of BERT and RoBERTa, AMDA achieves significant robustness gains under two strong adversarial attacks and alleviates the performance degradation of ADA on the clean data. Our code is available at: \url{https://github.com/thunlp/MixADA}.
\end{abstract}

\section{Introduction}

{\let\thefootnote\relax\footnotetext{$^*$ Equal contribution}}
{\let\thefootnote\relax\footnotetext{$^\dagger$ Corresponding author email: liuzy@tsinghua.edu.cn}}

Pretrained language models (PLMs) have established state-of-the-art results on various NLP tasks~\cite{BERT,RoBERTa,ALBERT} 
and the pretraining-then-finetuning paradigm has become the status quo. However, recent works have shown the adversarial vulnerabilities of PLMs, where PLMs finetuned on various downstream datasets are fooled by different types of adversarial attacks~\cite{TextFooler,SememePSO,BenchmarkMRC,BERT-ATTACK,BAE,T3}.

\begin{figure}[t]
\centering
\includegraphics[width=0.9\linewidth]{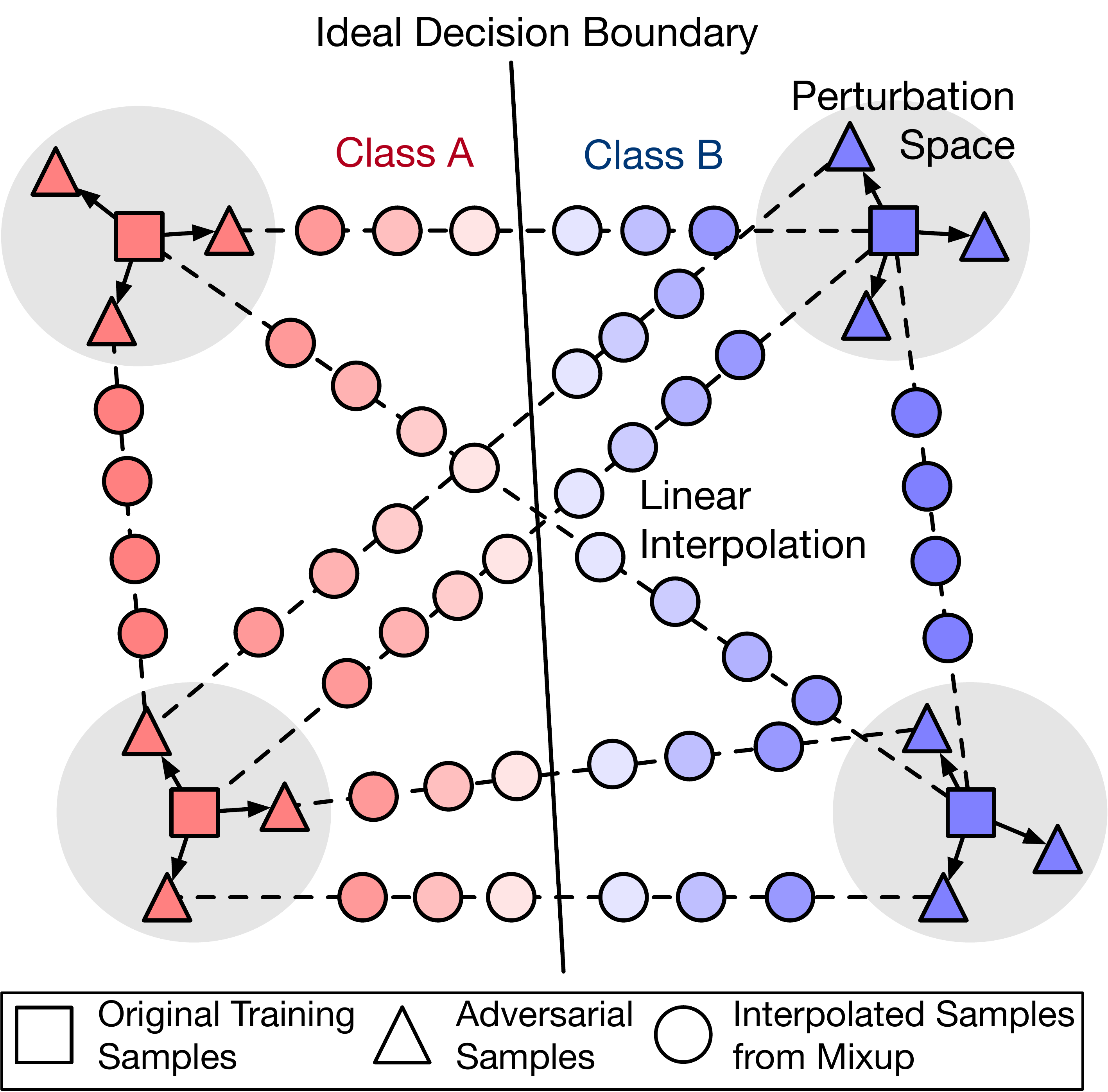}
\caption{Illustration of MixADA. Some of the interpolated samples are shown. We interpolate the representations of each pair of training samples including original samples and adversarial samples. Blue and red represent two different classes. The solid line represents the resultant decision boundary. AMDA helps achieve a more robust decision boundary.}
\label{fig:illustration}
\vspace{-1em}
\end{figure}

To improve adversarial robustness, two types of defense strategies have been proposed. The first type targets at specific attacks, such as spelling correction modules and pretraining tasks to defend character-level attacks~\cite{wordRec,robustEncoding,CharBERT} and certified robustness for word-substitution attacks~\cite{DeepMindIBP,robinIBP}. However, they are limited in practice as they are not generally applicable to other types of attacks. 
The other type of defense is Adversarial Data Augmentation (ADA), which augments the training set by the adversarial examples and is widely used in the training (finetuning) process to enhance model robustness~\cite{GA,PWWS,MHA,TextFooler,BERT-ATTACK,morphin,grammer-error,attackDependency,RL-NMT,CAT-Gen}.
ADA is generally applicable to any type of adversarial attacks but is not very effective in improving model performance under attacks. In this work, we aim to improve ADA and devise a general defense strategy to effectively improve model robustness during finetuning.\footnote{In this paper, we refer to such discrete adversarial training method as adversarial data augmentation to avoid confusion with the gradient-based adversarial training methods~\cite{AT-VAT}, which has been shown to be ineffective in defending against textual adversarial attacks~\cite{TAVAT}.} 

ADA has two major limitations for NLP models. Firstly, unlike images, it is harder to create new augmented textual data due to their discrete nature. Moreover, for textual adversarial attacks, the attack search space is prohibitively large. For example, the search space of word-substitution attacks consists of all combinations of the synonym replacement candidates, which is exponentially large. 
Consequently the number of adversarial training examples for augmentation is very insufficient. 
Secondly, ADA usually causes significant performance degradation on the clean data because the distribution of adversarial examples is very different from that of the clean data~\cite{PWWS}. 

In order to solve these two limitations, we create additional training samples via interpolating existing samples (Figure~\ref{fig:illustration}).
How to interpolate discrete textual inputs is non-trivial. We propose to convert the discrete textual inputs into continuous representations and then perform both ADA and mixup augmentation~\cite{mixup,NLPMix}, which is an augmentation technique proven to be particularly effective on continuous image data~\cite{IAT,MixInfer}.
We name our method Adversarial and Mixup Data Augmentation (AMDA).
With AMDA, we can create a much larger number of augmented training samples that cannot be obtained via discrete perturbations on textual data.
Moreover, AMDA's interpolated virtual training samples are closer to the distribution of the original data, which alleviates the performance degradation problem of ADA.



We experiment AMDA on three text classification datasets under two strong adversarial attacks and find that AMDA achieves significant robustness gains in all cases, notably restoring RoBERTa after-attack accuracy from 6.35\% to 51.84\% on IMDB, outperforming all other baselines by large margins.
Moreover, we also examine the evaluation method for adversarial robustness. Specifically, we find that the widely adopted \textit{Static Attack Evaluation} where a fixed set of adversarial examples are used to test all models is not reliable. In order to test model robustness under targeted attacks (i.e., not model-agnostic), we adopt the more challenging \textit{Targeted Attack Evaluation} where we generate a new set of targeted adversarial examples to evaluate each model. 
We encourage future defense works to also adopt this more reliable and challenging evaluation setting.

\section{Method}
In AMDA, we first augment training samples with ADA and then perform mixup during model training, where mixup augmentation is applied on the ADA-augmented training set.

\subsection{Adversarial Data Augmentation}

Given a victim model $f_v$ and the original training instances $\mathbf{D}_{ori}=\{(\mathbf{x}_i,\mathbf{y}_i)\}_{i=1}^n$, we employ an attacker to construct label-preserving adversarial training instances $\mathbf{D}_{adv}=\{(\mathbf{x}'_i, \mathbf{y}_i)\}_{i=1}^n$ such that: instances originally correctly classified are now classified wrongly ($f_v(\mathbf{x}'_i) \neq f_v(\mathbf{x}_i)$). We then train the model on the augmented training data $\mathbf{D}_{ADA}= \mathbf{D}_{ori} \cup \mathbf{D}_{adv}$.  

\subsection{Mixup Data Augmentation}



To better defend against the large number of possible adversarial examples, we propose to perform additional mixup augmentation during training.
Specifically, we linearly interpolate the representations and labels of pairs of training samples to create different virtual training samples, which can be formulated as:

\begin{equation}
\begin{aligned}
    \hat{\mathbf{x}} &= \lambda \mathbf{x}_i + (1 - \lambda ) \mathbf{x}_j, \\
    \hat{\mathbf{y}} &= \lambda \mathbf{y}_i + (1 - \lambda ) \mathbf{y}_j,
\end{aligned}
\end{equation}

where $(\mathbf{x}_i, \mathbf{y}_i)$ and $(\mathbf{x}_j, \mathbf{y}_j)$ are two labeled examples, and $\lambda \in [0,1]$ comes from a beta distribution $\lambda \sim Beta(\alpha, \alpha)$, where $\alpha$ is a hyper-parameter. On textual data, we cannot directly mix the discrete tokens. Instead, we can either interpolate the word embedding vectors or models' hidden representations of textual inputs. Meanwhile, we directly interpolate the labels, which are represented as one-hot vectors. 

When applied together with adversarial data augmentation, we allow the mixing of different types of data (between original examples, between original examples and adversarial examples, and between adversarial examples) to increase diversity. 


\begin{table*}[ht]
\centering
\small
\begin{tabular}{ l|rrrrr|rrrrr } 
 \toprule
 & \multicolumn{5}{c|}{SST-2} & \multicolumn{5}{c}{IMDB} \\
 & Original & PWWS-d & PWWS-s & TF-d & TF-s  & Original & PWWS-d & PWWS-s & TF-d & TF-s \\
 \midrule
 BERT$_{v}$ & 92.04 & 19.17 & 19.17 & 5.66 & 5.66 & 97.34 & 23.36 & 23.36 & 3.48 & 3.48 \\
 BERT$_{r1}$ & 91.10 & 18.73 & 44.98 & 3.46 & 45.36 & 96.72 & 25.61 & 69.88 & 1.64 & 76.64 \\
 BERT$_{r2}$ & 90.94 & 20.26 & 45.63 & 2.97 & 45.80 & 97.13 & 30.12 & 65.78 & 2.46 & 76.23 \\
 \midrule
 RoBERTa$_{v}$ & 94.45 & 25.48 & 25.48 & 3.29 & 3.29 & 97.75 & 15.98 & 15.98 & 1.84 & 1.84 \\
 RoBERTa$_{r1}$ & 94.29 & 31.03 & 50.47 & 5.82 & 41.63 & 97.54 & 27.46 & 65.57 & 3.48 & 77.46 \\
 RoBERTa$_{r2}$ & 93.85 & 32.13 & 50.69 & 9.34 & 40.91 & 97.34 & 17.21 & 73.36 & 2.87 & 76.64 \\
\bottomrule
\end{tabular}
\caption{Comparison between dynamic and static evaluation. PWWS-d, PWWS-s, TF-d, TF-s represent PWWS dynamic, PWWS static, TextFooler dynamic, TextFooler static, respectively. Numbers in the table represent accuracy. BERT$_{v}$ and RoBERTa$_{v}$ are the victim model for generating static evaluation examples. BERT$_{r1}$, BERT$_{r2}$, RoBERTa$_{r1}$, and RoBERTa$_{r2}$ are the fine-tuned models with new random seeds.}
\label{table:static}
\end{table*}

\subsection{AMDA}
In our proposed Adversarial and Mixup Data Augmentation (AMDA), 
we train the new model $f$ on the augmented training data $\mathbf{D}_{AMDA}$, which is obtained by performing both adversarial data augmentation and mixup data augmentation. We minimize the sum of the standard training loss and the mixup loss: 
\begin{equation}
\small
L = \sum_{i=1}^n L_{CE}(f(\mathbf{x}_i), \mathbf{y}_i) + \sum_{i=1}^m L_{KL}(f(\hat{\mathbf{x}}_i), \hat{\mathbf{y}}_i),
\end{equation}
where $(\mathbf{x}_i,\mathbf{y}_i)$ is from $\mathbf{D}_{ADA}$ and $(\hat{\mathbf{x}}, \hat{\mathbf{y}})$ is the virtual example obtained by applying mixup on the random pair of training data sampled from $\mathbf{D}_{ADA}$. We use cross-entropy to compute loss on $(\mathbf{x}_i,\mathbf{y}_i)$ and use KL-divergence for loss on $(\hat{\mathbf{x}}_i, \hat{\mathbf{y}}_i)$.

\section{Robustness Evaluation}


There are two different ways of robustness evaluation under adversarial attacks used in previous works. In this work, we explicitly differentiate them as Static Attack Evaluation (SAE) and Targeted Attack Evaluation (TAE):

\textbf{SAE} generates a fixed set of adversarial examples on the original model as the victim model.
This fixed adversarial test set will then be used to evaluate all the new models. This evaluation setup has been adopted in \cite[\textit{inter alia.}]{PWWS,morphin,grammer-error,CAT-Gen,RL-NMT,InfoBERT}.

\textbf{TAE} 
re-generates a new set of adversarial examples to target every model being evaluated. This is adopted in \cite[\textit{inter alia.}]{MHA,DeepMindIBP,robinIBP,BERT-ATTACK,SememePSO,attackDependency,TAVAT}


We observe that some authors did not explicitly specify the mode of evaluation in their papers\footnote{We had to email some of the authors to clarify the evaluation setup being adopted.}, leading to confusion and even conflicting conclusions. Thus, we explicitly differentiate the two modes of evaluation and provide a comparison in our experiments.

\begin{table*}[ht]
\centering
\small
\setlength{\tabcolsep}{1mm}{
\begin{tabular}{ l|cccr|cccr } 
 \toprule
 & \multicolumn{4}{c|}{SST-2} & \multicolumn{4}{c}{IMDB} \\
 & \multicolumn{2}{c}{PWWS} & \multicolumn{2}{c|}{TextFooler} & \multicolumn{2}{c}{PWWS} & \multicolumn{2}{c}{TextFooler} \\
 & Original & Adversarial & Original & \multicolumn{1}{c|}{Adversarial} & Original & Adversarial & Original & \multicolumn{1}{c}{Adversarial} \\
 \midrule
BERT & 91.27 & 14.83 \textit{(20.88\%)} & 91.27 & 2.97 \textit{(16.21\%)} & 97.75 & 24.18 \textit{(24.10\%)} & 97.75 & 1.64 \textit{(10.18\%)} \\
\quad+ADA &  90.12 & 27.18 \textit{(24.46\%)} & 90.50 & 9.01 \textit{(18.32\%)} & 96.93 & 25.82 \textit{(34.53\%)} & 96.93 & 3.07 \textit{(11.81\%)} \\
 \quad+TMix & 91.82 & 21.20 \textit{(19.36\%)} & 91.82 & 3.51 \textit{(16.39\%)} & 97.13 & 43.24 \textit{(32.51\%)} & 97.13 & 0.00 \textit{(12.06\%)} \\
\quad+SMix & 91.82 & 22.52 \textit{(20.47\%)} & 91.82 & 4.61 \textit{(16.76\%)} & 97.13 & 31.97 \textit{(23.74\%)} & 97.13 & 2.66 \textit{(12.39\%)} \\
\quad+AMDA-TMix & 91.54 & \textbf{38.82} \textit{(23.73\%)} & 91.93 & \underline{13.23} \textit{(19.66\%)} & 97.34 & \underline{51.02} \textit{(36.76\%)} & 96.72 & \underline{4.51} \textit{(17.23\%)} \\
\quad+AMDA-SMix & 91.10 & \underline{31.52} \textit{(24.11\%)} & 92.15 & \textbf{17.35} \textit{(18.64\%)} &  96.72 & \textbf{60.86} \textit{(27.79\%)} & 96.72 & \textbf{17.42} \textit{(13.85\%)} \\
\midrule
RoBERTa & 94.62 & 28.39 \textit{(23.06\%)} & 94.62 & 5.44 \textit{(18.51\%)} &  97.54 & 28.07 \textit{(37.48\%)} & 97.54 & 6.35 \textit{(12.61\%)} \\
\quad+ADA & 94.07 & 25.26 \textit{(27.07\%)} & 92.75 & 9.67 \textit{(19.71\%)} &  97.54 & 24.80 \textit{(49.36\%)} & 96.93 & 12.50 \textit{(14.39\%)} \\
\quad+TMix & 94.18 & 30.04 \textit{(23.19\%)} & 94.18 & 11.04 \textit{(17.69\%)} & 97.54 & 44.06 \textit{(39.33\%)} & 97.54 & 21.11 \textit{(14.01\%)} \\
\quad+SMix & 93.96 & 31.52 \textit{(22.86\%)} & 93.96 & 8.29 \textit{(17.80\%)} &  97.34 & 41.39 \textit{(34.90\%)} & 97.34 & 22.34 \textit{(11.96\%)} \\
\quad+AMDA-TMix & 93.90 & \underline{36.74} \textit{(26.02\%)} & 93.03 & \underline{13.78} \textit{(20.15\%)} &  98.57 & \underline{50.41} \textit{(59.68\%)} & 97.13 & \textbf{51.84} \textit{(16.62\%)} \\
\quad+AMDA-SMix & 93.96 & \textbf{41.85} \textit{(27.17\%)} & 93.47 & \textbf{16.80} \textit{(21.88\%)} &  97.54 & \textbf{55.12} \textit{(45.30\%)} & 97.54 & \underline{49.18} \textit{(15.52\%)} \\

\bottomrule
\end{tabular}}
\caption{Accuracy of the various models under PWWS and TextFooler attacks. Best performance for BERT-based models and RoBERTa-based models under each attack is \textbf{boldfaced}, the second best performance is \underline{underlined}. \textit{Numbers} in brackets indicate the average word modification rate of each attack.}
\label{table:results}
\end{table*}

\begin{table}[h]
\centering
\small
\setlength{\tabcolsep}{2.5mm}{
\begin{tabular}{ l|cccr } 
 \toprule
 & \multicolumn{2}{c}{PWWS} & \multicolumn{2}{c}{TextFooler} \\
 & Orig. & Adv. & Orig. & Adv.  \\
 \midrule
RoBERTa & 94.34 & 47.50 & 94.34 & 25.53 \\
\quad+ADA & 93.55 & 66.97 & 94.08 & 44.61 \\
\quad+TMix & 94.08 & 45.66 & 94.08 & 26.58 \\
\quad+SMix & 94.08 & 45.13 & 94.08 & 22.63 \\
\quad+AMDA-TMix & 94.47 & \underline{69.74} & 93.95 & \textbf{56.32} \\
\quad+AMDA-SMix & 94.34 & \textbf{70.00} & 93.42 & \underline{51.32} \\
\bottomrule
\end{tabular}}
\caption{Results on AG News multi-class classification dataset, with RoBERTa model.
Best performance under each attack is \textbf{boldfaced}, the second best performance is \underline{underlined}. }
\label{table:ag_news_results}
\end{table}

\section{Experiments}

\subsection{Experiment Setups}

\textbf{Datasets.} We evaluate our methods on three text classification datasets: two sentiment analysis datasets: SST-2~\cite{SST-2} and IMDB~\cite{IMDB}, where both datasets are binary classification tasks; as well as a multi-class news classification dataset AGNews~\cite{AGNews}, which consists of four different classes. For SST-2, we attack the entire test set (1821 samples) and report the accuracy under attacks. For IMDB, we find that it is prohibitively slow to attack the whole test set (25k samples) and hence we use the subset of the original test set as released in~\citet{contrastSet} for faster evaluation, which consists of 488 test instances. Similarly, on AGNews, we randomly sampled 10\% of the original test set and hold out as the test samples for attack evaluation. We also include these data splits in our released code base for easy reproduction and fair comparison for future works.

\textbf{Victim models and attack methods.} We experiment with both BERT-base-uncased~\cite{BERT} and RoBERTa-base~\cite{RoBERTa} as the victim models. We use PWWS~\cite{PWWS} and TextFooler~\cite{TextFooler} as our attack methods, which have been shown to effectively attack state-of-the-art NLP models including PLMs such as BERT. Both attack algorithms have access to model predictions but not gradients, and iteratively search for word synonym substitutes that flip model predictions without drastically changing the original semantic meanings and golden labels. 

\textbf{Details of mixup.} When performing mixup, we mix hidden representations of upper layers of BERT. The vectors used for mixup are hidden representations of the input examples at layer $i$ of the Transformer encoder, where $i$ is randomly sampled from $\{7,9,12\}$, which was found to be empirically effective~\cite{MixText}. Furthermore, we explore two different ways of obtaining the hidden representations of input examples from PLMs like BERT: (1) We use the vector of the \texttt{[CLS]} token at the $i$th-layer of BERT as the hidden representation for mixing. We name this approach \textbf{SMix}. (2) We perform mixup on every token's vector representation at the $i$th-layer. We name this approach \textbf{TMix}, which is the approach taken by \citet{MixText}.

\textbf{Details of ADA and AMDA.}
For both ADA and AMDA, we generate and add the corresponding adversarial examples of PWWS and TextFooler into training. For comparison, we also experiment with mixup alone without adding the adversarial examples. In this case, the model would only interpolate pairs of original training examples. 
We perform a greedy hyper-parameter search for the amount of augmented adversarial training samples and mixup parameter $\alpha$ as described in the Appendix. We also report average word modification rates, which indicate the percentage of words being replaced for attacking. Higher word modification rates indicate that the model is harder to attack and hence needs more words to be replaced.

\subsection{Comparison of SAE and TAE}

To compare SAE and TAE, we attack the finetuned model (BERT$_{v}$), RoBERTa$_{v}$ as the victim on SST-2 and IMDB, and then use the generated adversarial test set as the fixed test set for SAE. We then change the random seeds and re-finetune the models on the same data (BERT$_{r1}$, BERT$_{r2}$, RoBERTa$_{r1}$, RoBERTa$_{r2}$) with all other hyper-parameters being the same. We evaluate all these models using both SAE and TAE. The results are shown in Table~\ref{table:static}.

We find that by simply changing the random seeds, models achieve significant improvement under SAE. However, when we re-generate the adversarial test set for each model, their performances under TAE stay consistently poor. 
Moreover, we train BERT and RoBERTa with ADA and find that although BERT$_{\rm ADA}$ and RoBERTa$_{\rm ADA}$ perform well under SAE, they still perform poorly under TAE. This shows that conventional ADA is actually ineffective in improving model robustness under the challenging TAE setting.
We conclude that the adversarial examples found by the attackers target specifically at the victim models, hence they cannot fully reveal weaknesses of new models even if they only differ in random seeds.
We believe that TAE is the more challenging and meaningful evaluation method to measure model robustness under targeted attacks. 
We adopt TAE for the rest of the experiments in this paper and encourage future works to do so for fair comparison.

\subsection{Mixup Improves Robustness}


The comparison of AMDA and baseline methods under attacks for SST-2 and IMDB is shown in Table~\ref{table:results}. The results on the AGNews dataset with RoBERTa model is shown in Table~\ref{table:ag_news_results}. We observe that: 
(1) Mixup alone (both TMix and SMix) can often improve model robustness. For example, TMix and SMix improve the robust accuracy significantly under both attacks when using RoBERTa on IMDB. 
(2) AMDA (both AMDA-TMix and AMDA-SMix) can achieve further robustness improvement as compared to ADA and mixup in all cases. 
This proves that mixup and ADA can complement each other to better improve model robustness under adversarial attacks.
(3) Compared to ADA, our AMDA method does not incur significant performance degradation on the original test sets while improving robustness. In some cases, for example, BERT+TMix and BERT+AMDA-TMix even improve the model performance on the original test sets. This benefit is likely because that mixup creates virtual examples that are closer to the empirical data distribution.
(4) We find that models trained with AMDA also incur higher word modification rates under both attacks. For example, RoBERTa+AMDA-TMix incurs 59.68\% word modification rate under PWWS attack, while the RoBERTa baseline only needs 37.48\% words to be replaced. This further demonstrates that our proposed method improves robustness.

\section{Conclusion}

In this work, we propose AMDA as a generally applicable defense strategy by combining both adversarial and mixup data augmentation to cover more of the attack space. We show that AMDA greatly improves PLMs' robustness under the challenging TAE evaluation setting under
two strong adversarial attacks. 
We leave a more thorough theoretical analysis of AMDA's effectiveness on textual data as future work.\footnote{\citet{Zhang2020HowDM} provided a theoretical proof of mixup's effectiveness on continuous image data, which may serve as a foundation for more theoretical work.}
We believe that our work can establish the appropriate evaluation protocol and offer a competitive baseline for future works on improving the robustness of PLMs. 


\section*{Acknowledgments}
We thank members of THUNLP for their helpful discussion and valuable feedback on our work.
This work is supported by the National Key Research and Development Program of China (No. 2020AAA0106501) and Beijing Academy of Artificial Intelligence (BAAI).


\bibliography{acl2020}
\bibliographystyle{acl_natbib}

\clearpage
\appendix
\section*{Appendix}
\label{sec:appendix}

\section*{Hyper-parameter Analysis}

In this section, we perform further analysis to examine the effects of different hyper-parameters. 
There are two hyper-parameters involved in MixADA: the amount of adversarial data added for training, and the $\alpha$ parameter in the beta distribution of mixup coefficient.
We also experiment with an alternative ADA strategy - iterative ADA. 


\subsection*{Amount of Adversarial Training Data}
\begin{figure}[t]
\centering
\includegraphics[scale=0.38]{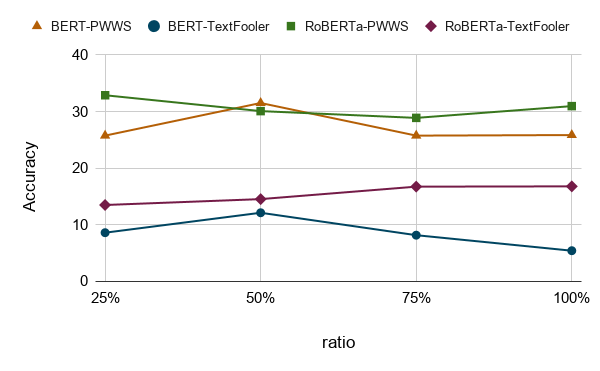}
\caption{Performance under attacks on the SST-2 dataset with varying ratio of adversarial training samples.}
\label{fig:ratios}
\end{figure}

We vary the ratio of the training dataset that we generate adversarial training samples on and add to the MixADA fine-tuning. We experiment with SMixADA with the hyper-parameter of mixup being fixed. On SST-2, we vary the ratio in $\{25\%, 50\%, 75\%, 100\%\}$. On IMDB, since the average sequence length is significantly longer and the adversarial example generation process becomes much slower, we experiment with a set of smaller ratios: $\{5\%, 10\%, 15\%, 20\%\}$. The results are plotted in in Figure~\ref{fig:ratios}. Interestingly, we find that higher ratio of adversarial training samples does not necessarily bring in additional robustness gains.

\subsection*{Interpolation Coefficient in Mixup}
\label{sec:hyperparameters}
\begin{figure}[t]
\centering
\includegraphics[scale=0.38]{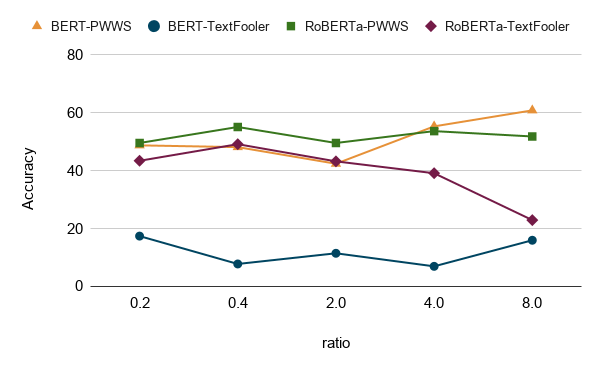}
\caption{Performance under attacks on the IMDB dataset with varying $\alpha$ parameter for mixup.}
\label{fig:alphas}
\end{figure}

We also analyse the hyper-parameter of mixup: the $\alpha$ parameter in the beta distribution, from which the interpolation coefficient is sampled. We fix the ratio of adversarial training data and vary $\alpha$ in the range of $\{0.2, 0.4, 2.0, 4.0, 8.0\}$. The results are plotted in Figure~\ref{fig:alphas}. We find that there is no consistent pattern across different datasets on what is the optimal $\alpha$. Hence, for our main experiments in the paper, we perform a greedy hyper-parameter search: we first tune the ratio of adversarial training samples, then fix the ratio and tune the $\alpha$ parameter for mixup. A more exhaustive hyper-parameter search might bring additional performance gains but would also incur extra computation costs.

\subsection*{Iterative ADA}

For our MixADA experiments in the paper, we generate all adversarial training samples at one shot and mix them with the original examples before fine-tuning. An alternative is to generate a new batch of adversarial training samples dynamically with the current model at each epoch. We compare this iterative approach with our MixADA and use the same ratio of adversarial training samples and mixup parameter $\alpha$. We evaluate RoBERTa on the SST-2 dataset. The results are in Table~\ref{table:iterative}.

\begin{table}[h]
\centering
\small
\begin{tabular}{ lrr } 
 \toprule
  & PWWS & TextFooler \\
\midrule
  TMixADA & 36.74 & 13.78 \\
\quad+iterative & 28.45 & 6.26 \\
SMixADA & 41.85 & 16.80 \\
\quad+iterative & 28.78 & 7.69 \\
\bottomrule
\end{tabular}
\caption{Performance of MixADA under attacks in the one-shot approach and the iterative approach.}
\label{table:iterative}
\end{table}

We find that the iterative approach is far worse than our one-shot approach. We hypothesize that in the one-shot approach, we generate the adversarial examples on a fully-fine-tuned model while the iterative approach generates adversarial examples on the not-well-fine-tuned model in the first few epochs, and hence the adversarial examples generated in the iterative approach are not as challenging and useful as those in our one-shot approach.

\end{document}